\documentclass[conference]{IEEEtran}
\usepackage[T1]{fontenc} % optional
\usepackage{amsmath}
\usepackage{cite}
\usepackage{amsfonts}
\usepackage{amssymb}
\usepackage{graphicx}
\usepackage{textcomp}
\usepackage{xcolor}
\usepackage{microtype}
\usepackage{subfigure}
\usepackage{ctable}
\usepackage{booktabs}
\usepackage{algpseudocode}
\usepackage{algorithm}
\usepackage{hyperref}
\begin{document}

\title{TracInAD: Measuring Influence for Anomaly Detection\thanks{Accepted at the Proceedings of IJCNN 2022.}\thanks{© 2022 IEEE. Personal use of this material is permitted. Permission from IEEE must be
obtained for all other uses, in any current or future media, including
reprinting/republishing this material for advertising or promotional purposes, creating new
collective works, for resale or redistribution to servers or lists, or reuse of any copyrighted
component of this work in other works.}}
%should not be used}}
%}

\author{\IEEEauthorblockN{Hugo Thimonier\IEEEauthorrefmark{1}, Fabrice Popineau\IEEEauthorrefmark{1}, Arpad Rimmel\IEEEauthorrefmark{1}, Bich-Li\^en Doan\IEEEauthorrefmark{1} and Fabrice Daniel\IEEEauthorrefmark{2}}
\IEEEauthorblockA{\IEEEauthorrefmark{1} Université Paris-Saclay, CNRS, CentraleSupélec, Laboratoire Interdisciplinaire des Sciences du Numérique, \\ 91190, Gif-sur-Yvette, France. \\
Email: name.surname@lisn.fr}
\IEEEauthorblockA{\IEEEauthorrefmark{2}\textit{Artificial Intelligence Department of Lusis},
Paris, France.}}

\maketitle

\begin{abstract}
    As with many other tasks, neural networks prove very effective for anomaly detection purposes. However, very few deep-learning models are suited for detecting anomalies on tabular datasets. This paper proposes a novel methodology to flag anomalies based on TracIn, an influence measure initially introduced for explicability purposes. The proposed methods can serve to augment any unsupervised deep anomaly detection method. We test our approach using Variational Autoencoders and show that the average influence of a subsample of training points on a test point can serve as a proxy for abnormality. Our model proves to be competitive in comparison with state-of-the-art approaches: it achieves comparable or better performance in terms of detection accuracy on medical and cyber-security tabular benchmark data.
\end{abstract}

\section{Introduction}
\label{introduction}
     
    As a research direction, anomaly detection has caught more and more attention in recent years due to its many successful applications across a large number of domains. This field of research bears several names that point towards relatively similar methods and approaches \cite{ruffUnifyingReviewDeep2021a}: outlier, novelty and anomaly detection. The difference in the pursued objectives can account for the dissimilarity in semantics.
    
    Scholars often consider its utility to be two-fold. On the one hand, it is well-suited for fraud and intrusion detection problems that involve highly imbalanced datasets for which standard supervised learning methods often fail. Most of these supervised approaches fail when over-sampling and under-sampling cannot be efficiently applied, which is often the case for tabular datasets. On the other hand, anomaly detection has also proven to be very effective when very few or no labels are available.

    In short, scholars have defined anomaly detection as the process of identifying points that deviate from a pre-defined notion of normality. Common examples of anomaly detection include flagging frauds among standard credit card payments, identifying mislabeled samples within a dataset or detecting computer intrusion. 
    
    The literature investigates two approaches to anomaly detection. The traditional supervised approach involves giving classifiers anomalous and normal samples as inputs in the training phase. A somewhat more hegemonic approach to anomaly detection involves learning or characterizing a distribution during training by considering a training set only composed of normal data. Anomalies are then identified in the inference stage by evaluating how well each sample fits the estimated distribution.

    Many researchers have recently proposed different methods to try and tackle the critical problem of detecting anomalies. Most of today's state-of-the-art methods rely on deep learning models such as Deep SVDD \cite{deep-svdd}, which mimic the canonical approach of SVDD \cite{svdd} without using the computationally costly kernels. More recently, methods that rely on self-supervision, such as \cite{goad} which uses pretext tasks, or \cite{neutralad} which involves a contrastive loss, have also proven to perform well. Except for \cite{goad} and \cite{neutralad}, most proposed methods focus on applications related to image or textual datasets.

    This paper introduces a novel approach to anomaly detection based on influence measures. Measuring influence here describes the task of evaluating how much a sample contributed to increasing or decreasing the loss of another sample in the course of training. Our methodology relies on $\texttt{TracIn}$ proposed in \cite{tracin}, to evaluate the average influence of a subsample of training points on a sample. Following up on \cite{tracinvae} who showed that self-influence in the unsupervised set-up is correlated with the loss of the sample, we propose a standalone method based on Variational Autoencoders \cite{kingma2014autoencoding} which outperforms or shows comparable results with state-of-the-art anomaly detection methods. 

    The main contributions involved in our methodology are:
    \begin{itemize}
        \item A novel anomaly detection method based on Variational Autoencoders and influence measures which offers competitive results on several benchmark datasets.
        \item We display evidence that the proposed approach can be extended to any deep anomaly detection method.
    \end{itemize}

\section{Related Works}
    \label{relatedworks}
    \subsection{Anomaly Detection}
        Anomaly detection can be divided into the following non-exhaustive categories:

        \textbf{One-Class Classification}. One-Class Classification involves discriminative models for anomaly detection: such methods avoid the challenge of estimating the normal distribution as a means of identifying anomalies. Instead, those methods aim at characterizing the density by proposing a decision boundary. In short, One-Class Classification describes methods that use exclusively normal data in the training stage to learn a decision function to flag anomalies. This sub-field includes both shallow and deep anomaly detection methods. For instance, in OCSVM \cite{ocsvm}, and SVDD \cite{svdd}, authors propose to learn a decision function in the form of a hyperplane and a hypersphere respectively, in a Hilbert space by using the kernel trick. Regarding deep methods, \cite{deep-svdd}, \cite{deep-semi-svdd} put forward Deep-SVDD in which a deep neural network replaces kernels to map data points to a latent space so that normal points will be contained in a hypersphere. Deep-SVDD can suffer from \textit{model collapse}, which designates a trivial mapping to a unique point in the latent space for certain network architectures. Thus, in recent studies \cite{goyalDROCCDeepRobust2020}, \cite{chongSimpleEffectivePrevention2020} have proposed regularization methods to avoid such pitfalls. 

        \textbf{Reconstruction based methods}. Reconstruction based methods focus on learning a model which reconstructs well normal instances while failing to reconstruct abnormal points. Thus, the reconstruction error can serve as a proxy for anomaly: the higher the error, the higher the probability of a point being an anomaly. Such reconstruction-based methods encompass deterministic methods such as autoencoders, PCA, or probabilistic variants. For instance, PCA has been adapted to anomaly detection in \cite{pca_matrix_sketching} or \cite{pca}. Other methods have relied on autoencoders to estimate the normal data distribution, such as \cite{Kim2020RaPP} in which authors augment the reconstruction error with differences in the latent representations of the original and reconstructed samples. 

        \textbf{Self-supervised methods}. In a large spectrum of applications including anomaly detection, many methods now involve self-supervision as a means of improving models' learning capacity. For instance, in \cite{goad} authors propose to transform data samples and use the self-supervised pretext task of identifying which transformation was applied to a sample. Failure to correctly predict the applied transformation can serve as a proxy measuring anomaly. In a similar approach, \cite{neutralad} authors also rely on transformations to flag anomalies but propose to learn them instead of using affine transformation as in \cite{goad}. Another recent approach, \cite{shenkar2022anomaly}, fetches internal contrastive learning to flag anomalies on tabular data. Other recent work in \cite{sohn2021learning} proposed a methodology for contrastive pretraining to improve on anomaly detection methods through a two-staged framework. 

    \subsection{Influence Estimation}
        Related to our work, influence estimation has also attracted growing interest from the research community. It designates the identification of the training samples most responsible for the prediction of a particular test sample $x$. This involves the computation of an influence score which can take many different forms \cite{pmlr-v108-barshan20a}, \cite{pmlr-v119-basu20b}, \cite{tracin} \cite{yehRepresenterPointSelection}. Recently, \cite{pmlr-v70-koh17a} proposed the influence function, which relies on the idea that if an influential sample for a particular test sample is removed from the set used for training, then the test sample's loss should significantly increase. Since estimating such function may be hard, in \cite{tracin} authors propose $\texttt{TracIn}$ a computationally efficient first-order approximation to the influence function.
        As mentioned by the authors, such an influence function can be used to flag odd points in a dataset: mislabeled samples, outliers or even anomalies. For instance, in an unsupervised set-up, where they use $\beta$-Variational Autoencoders, \cite{tracinvae} analyze the behaviour of the influence function. They show that, for points that stand out from the rest of the dataset, self-influence, which measures the influence of a sample on its own loss/likelihood, tends to be larger than for normal points. Based on their diagnostic, we hypothesize that not only do self-influence behaviours differ between normal and abnormal points, but the influence of normal points on abnormal points should significantly differ from the influence of normal points on normal points.

        In this work, we propose using influence measures to flag anomalies among normal samples. Firstly, we propose a standalone method based on a Variational Autoencoder, which shows that alone influence can be used efficiently for detecting anomalies. Secondly, we discuss that self-influence can serve to augment anomaly scores in many deep anomaly detection methods.

\section{Method}
    
    \subsection{$\texttt{TracIn}$}
        \label{tracin_subsection}

        Let us briefly discuss $\texttt{TracIn}$ \cite{tracin} as a means of measuring influence. 
        Consider the following set up where $\mathcal{X}\subseteq \mathbb{R}^d$ represents the data space, and a dataset $\mathcal{D}_n = \{x_j\}_{j=1}^n, x_j \in \mathcal{X}$. Let $f_\theta$ denote a model, parametrized by $\theta \in \Theta \subseteq \mathbb{R}^p$, whose parameters are obtained through optimizing a loss function $\ell:\Theta \times \mathcal{X} \rightarrow \mathbb{R}$. The loss of the model parametrized by $\theta$ for a data point $x \in \mathcal{X}$ is $\ell(\theta, x)$. 
        For a training set $\mathcal{D}_{train} \subseteq \mathcal{D}_n$, the parameters of the model are obtained through minimizing $\sum_{x \in \mathcal{D}_{train}} \ell(\theta, x)$.
        
        Define the influence of a sample $x$ on a test sample $x'$ as the difference in the loss for the sample $x'$ incurred by having included $x$ in the training set. Formally, the influence function of a sample $x$ on the test sample $x'$ is:
        \begin{equation}
        \label{influence_function}
        IF(x, x') = \ell(\theta, x') - \ell(\theta_{-x}, x')
        \end{equation}
        where $\theta_{-x}= \arg \min_{\theta \in \Theta} \sum_{z \in \mathcal{D}_{train} \setminus\{x\}} \ell(\theta, z)$.
        
        Consider an iterative optimization process, \textit{e.g.} Stochastic Gradient Descent (SGD), where $\theta_t$ denotes the obtained parameters after iteration $t$, $B_t$ a minibatch of size $b$ at iteration $t$, and a step size $\eta_t$ at iteration $t$. $\texttt{TracIn}$ gives the following measure of influence of $x$ on the test sample $x'$ when SGD is the optimizer\footnote{We refer the reader to \cite{tracin} for more detail on how \eqref{tracin_eq} is derived.}:
        \begin{equation}
        \label{tracin_eq}
            \texttt{TracIn}(x,x') = \frac{1}{b} \sum_{t: x \in B_t} \eta_t \nabla\ell(\theta_t,x') \cdot \nabla\ell(\theta_t,x)
        \end{equation}
        where $\nabla\ell(\theta_t,x')$ denotes the gradient of the loss function evaluated for the sample $x'$ w.r.t. the parameter $\theta_t$.
        
        To avoid excessive computational overhead, one uses $\texttt{TracInCP}$ given in \eqref{tracincp}. It consists in storing checkpoints (CP) along the training process, \textit{i.e.} parameters $\theta_{t_1}, \theta_{t_2}, \hdots, \theta_{t_k}$ corresponding to iteration $t_1, t_2, ..., t_k$ assuming that between checkpoints each training sample is visited only once, \textit{e.g.} one epoch, and that the step size remains constant between checkpoints.
        \begin{equation}
        \label{tracincp}
            \texttt{TracInCP}(x,x') = \sum_{i=1}^k \eta_i \nabla\ell(\theta_{t_i},x') \cdot \nabla\ell(\theta_{t_i},x)
        \end{equation}
        
    \subsection{Influence as an anomaly score: $\texttt{TracInAD}$}
    \label{tracinad_}
    Consider a deep anomaly detection model $f_\theta$ whose parameters were optimized through minimizing a loss function $\ell(.,\theta)$ as detailed in section \ref{tracin_subsection}, and a training set  $\mathcal{D}_{train}$ solely composed of normal samples.
    
    Under the hypothesis that the normal samples belonging to the training set and the validation set were obtained from a similar distribution $p_{normal}$, the overall influence of training samples on normal validation samples should be positive. In other words, training samples should mostly reduce the loss of normal samples. Using the terminology proposed in \cite{tracin}, training samples should mostly be strong \textit{proponents} for any normal sample in the validation set. On the other hand, since anomalies' distributions differ from $p_{normal}$, training points should mostly be \textit{opponents}, in the sense that they contributed to increase the loss, or weak \textit{proponents} to the anomalies in the test set.
    
    Based on these premises, we propose to derive an anomaly score based on $\texttt{TracInCP}$. Consider a sample $x' \in \mathcal{D}_{val}$ for which we wish to assess whether it is an anomaly. At each checkpoint $t_i$, randomly select a subsample of the training set of fixed size $m$, denoted $B_{t} \subseteq \mathcal{D}_{train}$, and compute the average $\texttt{TracInCP}$ influence. This gives the following anomaly score:
    \begin{equation}
    \label{tracin_anomaly}
        \texttt{TracInAD}(x') =  \frac{1}{m} \sum_{x \in B_{t}} \sum_{i=1}^k \eta_i  \nabla\ell(\theta_{t_i},x') \cdot \nabla\ell(\theta_{t_i},x)
    \end{equation}
    Pseudo code of the $\texttt{TracInAD}$ algorithm is presented in Algorithm \ref{alg:tracinad}.
    
    \begin{algorithm}[t!]
        \begin{algorithmic}
        %\label{alg:tracinad}
        \caption{Pseudo Python Code for $\texttt{TracInAD}$}\label{alg:tracinad}
        \Require{$
        \mathcal{D}_{train}, \mathcal{D}_{val}, \{\theta_{t_1},\dots,\theta_{t_n}\},
        \{\eta_{t_1},\dots,\eta_{t_n}\},$ \newline 
        \hspace*{3em} $\ell(\theta,.), m$}
        \State $\texttt{TracInAD} \gets dict()$ 
        \State $B \gets$ random sample of size $m$ from $\mathcal{D}_{train}$
        \For{$t \in \{t_1,\dots,t_n\}$}
            \State $\theta \gets \theta_{t}$
            \State $\eta \gets \eta_{t}$
            \For{$x \in \mathcal{D}_{val}$}
                \State $\texttt{TracInAD}[x] \mathrel{{+}{=}} \frac{1}{m} \sum_{x' \in B} \eta \nabla \ell(\theta,x') \cdot \nabla\ell(\theta,x)$
            \EndFor
        \EndFor
        \end{algorithmic}
    \end{algorithm}

    \subsection{Application}
    \label{appl}
    To evaluate the pertinence of our methodology, we experiment using a vastly used deep anomaly detection method based on Variational Autoencoders (VAE) and the reconstruction error.
    
    \subsubsection{Variational Autoencoders}
    First proposed in \cite{kingma2014autoencoding}, Variational Autoencoders (VAE) are a particular form of autoencoders that rely on a Bayesian framework.
    Assume that every sample $x \in \mathcal{D}_{train}$ is obtained from some random process involving an unobserved continuous variable $z$. The latter is sampled from some prior distribution $p_{\psi^*}(z)$, while $x$ is obtained from a conditional distribution $p_{\psi^*}(x|z)$. In the VAE framework, the core objective is two-fold: (i) estimating the latent variable $z$ given a sample $x \in \mathcal{X}$ and (ii) estimating the parameter $\psi^*$ which fully describes the conditional and prior distributions. To do so, a recognition model $q_\phi(z|x)$, in the form of a neural network, is involved in the process to compensate for the intractability of the true posterior distribution $p_{\psi^*}(z|x)$. 
    Note that the recognition process $q_\phi(z|x)$ and the conditional distribution $p_{\psi^*}(x|z)$ are also referred to as, respectively, the encoder and the decoder.
    
    One seeks to find the parameter $\psi$ of the marginal distributions $p_\psi(x)$ such that the log-likelihood of the training set, $\sum_{x \in \mathcal{D}_{train}} \log p_\psi(x)$, is maximized. Conjointly, the recognition model is estimated such that it is as close as possible to the true posterior distribution. This objective is met through minimizing the Kullback-Leibler divergence between the two distributions, $D_{KL}(q_\phi(z|x)\|p_\psi(z|x))$. In the end, the VAE loss which is minimized w.r.t. $\{\psi, \phi\}$, can be expressed as 
    \begin{equation}
    \label{vae_loss}
    \begin{array}{rcl}
        \mathcal{L}_{VAE}(\psi, \phi) & = & - \mathbb{E}_{z \sim q_\phi(z|x)} \log p_\psi(x|z)\\
        && + D_{KL}(q_\phi(z|x)\|p_\psi(z|x)) \\
    \end{array}
    \end{equation}
    since $\log p_\psi(x) = \mathbb{E}_{z \sim q(z|x)} \log p_\psi(x|z)$. This loss can me minimized through backpropagation using the reparameterization trick \cite{kingma2014autoencoding}. 
    
    Given \eqref{vae_loss}, the influence function described in \eqref{influence_function} is expressed as follows:
    \begin{equation}
    \label{inf_vae}
    \begin{array}{@{}rcl}
        IF_{VAE}(x,x') & = &  - ( \mathbb{E}_{z \sim q_{\phi}(z|x')} \log p_\psi(x'|z) \\
        && - \mathbb{E}_{z \sim q_{\phi_{-x}}(z|x')} \log p_{\psi_{-x}}(x'|z)) \\
        && +  ( (D_{KL} (q_\phi(z|x')\|p_\psi(z|x') ) \\ 
        && - D_{KL} (q_{\phi_{-x}}(z|x')\|p_{\psi_{-x}}(z|x')))
    \end{array}
    \end{equation}
    
    In the present case, we consider a Gaussian prior $p_\psi(z) = \mathcal{N}(z; 0, I)$. Similarly, let the conditional distribution $p_\psi(x|z)$ be a multivariate Gaussian whose distribution parameters are computed from $z$ using a neural network. The variational approximate posterior is set to be a multivariate gaussian with a diagonal covariance matrix. Let a sample $x \in \mathcal{D}_{train}$, then
    $\log q_\phi(z|x) = \log \mathcal{N}(z; \mu(x), \sigma^2(x)\cdot I)$. 
    
    Empirically, a sample $x \in \mathcal{D}_{train}$ is fed to the encoder which outputs parameters $\{\mu(x), \sigma(x)\}$. Those parameters then allow sampling a latent vector $z$ corresponding to the original sample $x$. Then, the latent vector is fed to the decoder, which outputs a reconstructed sample $\tilde{x}$.
        
    \subsubsection{Reconstruction Error}
    \label{influence_ad}
    
    A standard approach to anomaly detection that relies on Autoencoders and VAE consists in using exclusively normal samples in the training phase to estimate the normal distribution. During the inference phase, the anomaly score is derived from the reconstruction error.
    Formally, consider a sample $x' \in \mathcal{D}_{val}$, and a trained autoencoder $f_{\theta^*}$ whose parameters $\theta^*$ were obtained through an iterative process described in section \ref{tracin_subsection}. Denote by $\tilde{x}' = f_{\theta^*}(x')$ the reconstructed sample $x'$ using the learned autoencoder. The anomaly score $s$ is obtained as follows
    \begin{equation}
    \label{recon_error}
    s(x') = \| x' - \tilde{x}' \|_2^2
    \end{equation}
    The described method relies on the same hypothesis as described in section \ref{tracinad_}. 
    \subsubsection{$\texttt{TracInAD}$ applied to VAE}
    Algorithm \ref{alg:tracinad} can also be applied to such set-up: $\texttt{TracInAD}$ can be used as an anomaly score instead of $s(x')$. Though, as seen in \eqref{inf_vae}, computing the exact influence function for a VAE can be challenging since it would involve computing an expectation over the encoder. However, \cite{tracinvae} have proven that under mild conditions, the empirical average of the influence function over $l$ \textit{i.i.d.} samples is close to the true influence function with high probability when $l$ is properly selected. Therefore, empirically we compute the average loss over $l$ reconstructed samples for each sample considered.
    
\section{Experiments}

    \subsection{Tabular Datasets}

        We propose to evaluate the performances of the model presented in \ref{appl} concerning anomaly detection in a tabular dataset set-up. We experiment on four benchmark datasets used in the literature. We follow the procedure first proposed in \cite{zong2018deep}.
    
        {\bf Arrhythmia dataset.} It consists of a cardiology dataset from the UCI repo: it contains attributes related to the diagnosis of cardiac arrhythmia in patients. The dataset is comprised of $16$ classes: class $1$ are normal patients, class $2$ to $15$ are arrhythmia suffering patients. The smallest classes $(3,4,5,7,8,9,14,15)$ are taken to be anomalous while the rest normal. Note that categorical attributes are dropped: in total, there are $274$ attributes. The train set is composed of 193 samples, while the validation set contains 259 samples.

        {\bf Thyroid Dataset.} It is another medical dataset from the UCI repo which contains attributes corresponding to whether a patient suffers from hyperthyroid. There are three classes in the dataset among which \textit{hyper-function} is designated as the anomalous class while the rest is normal. Only the $6$ continuous attributes are used. The train set and validation set are composed of $1,839$ and $1,933$ samples, respectively.

        {\bf KDD and KDDRev Datasets.} KDD is an intrusion dataset that was created by an extensive simulation of a US Air Force LAN Network. The dataset consists of a normal data class and $4$ simulated attack types (denial of service, unauthorized access from a remote machine, unauthorized access from local superuser and probing). $41$ different attributes are considered: $34$ are continuous, and $7$ are categorical. Categorical features are encoded using one-hot encoding.
        
        This dataset is used to construct two datasets. Firstly, the KDD dataset in which non-attack classes, which correspond to 20\% of the dataset, are treated as the anomaly class. Secondly, the KDD Reverse  dataset (KDDRev), in which the attack class is subsampled to consist of 25\%  of the number of non-attack samples and is considered the anomaly class. The train set of KDDRev (resp. KDD) is composed of $48,639$ (resp. $198,371$) samples, while the validation contains $72,958$ (resp. $295,650$) samples. 
        
        Following the configuration of \cite{zong2018deep}, each training set is composed of 50\% of the normal data. The 50\% remaining and the entirety of the anomaly samples compose the validation set. The share of anomalies in each validation set is the following: (i) Arrhythmia: $25.48$\% (ii) Thyroid: $4.8$\%, (iii) KDD: $32.9$\% (iv) KDDRev: $33.33$\%.

        \begin{table*}[ht!]
    \centering
    \begin{tabular}{ccccccccc}
        \specialrule{.1em}{.05em}{.05em} 
        Method & \multicolumn{8}{c}{Dataset} \\
        \hline 
        & \multicolumn{2}{c}{Arrhythmia} & \multicolumn{2}{c}{Thyroid} & \multicolumn{2}{c}{KDD} & \multicolumn{2}{c}{KDDRev} \\ \cline{2-5} \cline{6-9}
        & $F_1$ Score & $\sigma$ & $F_1$ Score & $\sigma$ & $F_1$ Score & $\sigma$ &$F_1$ Score & $\sigma$ \\
        \hline 
        OC-SVM & 45.8 &&  38.9&& 79.5 &&83.2& \\
        E2E-AE & 45.9 && 11.8&& 0.3 &&74.5& \\
        LOF & 50.0 & &52.7 & &83.8 & &81.6 & \\
        DAGMM \cite{zong2018deep}& 49.8 &&47.8&& 93.7&& 93.8& \\
        GOAD \cite{goad}& 52.0 & 2.3 & 74.5 & 1.1 & 98.4 & 0.2 & 98.9 & 0.3 \\
        NeuTraL AD \cite{neutralad}&  60.3 & 1.1 & 76.8 & 1.9 & 99.3 & 0.1 & 99.1 &  0.1\\
        Shenkar et al.\cite{shenkar2022anomaly} & \textbf{61.8} & 1.8 & 76.8 & 1.2 & \textbf{99.4} & 0.1 & \textbf{99.2} & 0.3 \\
        \hline 
        $\texttt{TracIn AD} $ & 54.6 & 2.1 & {\bf 77.6} & 5.4 & 82.1 & 0.6 & 98.8 & 0.3 \\
        \specialrule{.1em}{.05em}{.05em} 
    \end{tabular}
    \label{tab:tabular_res}
    \caption{Anomaly Detection Accuracy}
    \end{table*}

    \subsection{Hyperparameters}
        As discussed in section \ref{influence_ad}, we train a VAE for each of the datasets presented in the previous section exclusively on normal samples. For the smaller datasets, namely Arrhythmia and Thyroid, we consider an encoder and decoder comprised of 3 layers for the former dataset while 2 layers for the latter. Regarding the larger datasets, KDD and KDDRev, we resort to 6-layer networks for both the encoder and decoder. \newline
        We consider a batch size of 32 for the Arrhythmia dataset while 16 for Thyroid, the models were trained for 20 epochs and 250 epochs respectively. Note that the small size of both datasets involves a swift training process allowing a large number of epochs without any significant computational overhead. Regarding the KDD and KDDRev datasets, models were trained for 20 epochs, with a batch size of 32 and 128 respectively. We used the SGD algorithm to optimize the parameters of the network along the training process with learning rate set to $10^{-4}$ for both Thyroid and Arrhythmia datasets, while $10^{-5}$ for KDDRev and $10^{-7}$ for KDD. We refer the reader to the code made available online for more detail on the hyperparameter set-up\footnote{\href{https://github.com/hugothimonier/TracInAD}{https://github.com/hugothimonier/TracInAD}}.

        Regarding the computation of the anomaly score, we consider a step size between each saved checkpoint of 1 for Arrhythmia, 10 for Thyroid and 2 for both KDD and KDDRev. In other words, we compute the $\texttt{TracIn}$ measure of influence every 10 epochs, \textit{i.e.} for 25 CP in the case of Thyroid.
        The size $m$ of the batch containing random samples belonging to $\mathcal{D}_{train}$ used to compute $\texttt{TracInAD}$, as detailed in \ref{tracin_anomaly}, is set to 128 for Arrhythmia, 64 for Thyroid datasets while 256 for KDD and KDDRev. We set $l$ to 8 for the Thyroid dataset, 32 for Arrhythmia. Note that for faster training, we considered for both larger datasets $l=1$ and did not observe any deterioration of the performances in comparison with larger values.

    \subsection{Results}

    The results of the experiments can be observed in Table I. We compare our model to both shallow and deep anomaly detection methods. Note that results for models such as OCSVM, Local Outlier Factor (LOF), DAGMM \cite{zong2018deep} and GOAD \cite{goad} were directly taken from \cite{goad}, while results for NeuTraL AD \cite{neutralad} were taken from their paper. The F1-Score was chosen as the metric to compare the competing models in accordance with the literature.

    For our model, we performed 10 iterations for the smaller datasets and 5 for the larger datasets. The rows of the table indicate the model used while columns indicate obtained mean metric for the F1-Score over the different iterations (the higher, the better) as well as its standard deviation in the columns denoted by $\sigma$ for every dataset.
    
    We observe that our model outperforms all state of the art models on the Thyroid dataset. This may indicate that our model performs well on severe imbalanced set-ups: the Thyroid dataset stands out from the other three since it disposes of a much lower proportion of anomalies in the validation set. However, we also observe a relatively high standard deviation compared to other approaches. Our model competes well with \cite{shenkar2022anomaly}, NeuTraL AD \cite{neutralad} and GOAD \cite{goad} on the KDDRev dataset and outperforms all other considered approaches. Also, note that for the arrthythmia dataset our method performs on par in comparison with other methods while it fails to perform well on the KDD dataset.
    
\section{Discussion}

    \textbf{Generality of $\texttt{TracInAD}$}
    As discussed in section \ref{tracinad_}, measuring influence as an anomaly score can be used for any deep anomaly detection method. For instance, one can easily apply $\texttt{TracInAD}$ in the Deep SVDD \cite{deep-svdd} framework, where $\ell(., \theta)$ is taken to be the standard Deep-SVDD loss used to optimize the network's parameters.
    To support our statement regarding the generality of our approach, we experiment on the Thyroid dataset in the Deep SVDD framework. We train a 3-layer encoder by minimizing the Deep SVDD loss. Note here that no pre-training was performed to initialize the weights of the network. The model is trained for 50 epochs with batch size 16. The step used to compute the $\texttt{TracInAD}$ score is set to 10 as in the VAE set up. We compare the accuracy obtained with the standard Deep SVDD anomaly score, with the score computed as in \eqref{tracin_anomaly} where $\ell(\theta,x)$ is the standard Deep-SVDD loss used to optimize the parameters of the network.
    We obtain a mean F1-Score of $50.1\%$ for $\texttt{TracInAD}$-DSVDD over 200 iterations, in comparison with standard Deep-SVDD anomaly score, which obtains a F1-score of $40.1\%$. The difference is statistically significant. However, note that obtained results suffer from high standard deviation, which may mitigate the strength of our statement. 
    
    \textbf{High Standard Deviation} We observe that our model suffers from a higher standard deviation than competing models. We hypothesize that our model performances fluctuate more than other approaches mostly because it tends to exacerbate the original model's performance. On the one hand, when $f_\theta$ performs well, it tends to outperform the original model. On the other hand, when the original model performs poorly, $\texttt{TracInAD}$ appears to perform worst than the original model.
    
\section{Conclusion}

    We presented a methodology to detect anomalies based on a measure of influence which was first proposed in the explicability literature. Our methodology has the advantage of being adaptable to any deep anomaly detection and can improve on the standard anomaly score as discussed in the previous section.
    We also showed through an experiment using VAEs that our model competes well on tabular datasets, especially the most challenging ones, in comparison with the current state-of-the-art models. 

\section*{Acknowledgment}
This work was performed using HPC resources from the "M\'esocentre" computing center of CentraleSup\'elec and \'Ecole Normale Sup\'erieure Paris-Saclay supported by CNRS and Région Île-de-France (\href{http://mesocentre.centralesupelec.fr/}{http://mesocentre.centralesupelec.fr/}).
This research publication is supported by the Chair "Artificial intelligence applied to credit card fraud detection and automated trading" led by CentraleSupelec and sponsored by the LUSIS company.

\bibliographystyle{./bibliography/IEEEtrans.bst}
\bibliography{./bibliography/mybib.bib}

\end{document}